\documentclass[letterpaper,twocolumn]{article}
\usepackage{aaai16}
\usepackage{times}
\usepackage{helvet}
\usepackage{courier}
\usepackage{graphicx}
\usepackage{latexsym}
\usepackage{amsmath}
\usepackage{amssymb}
\usepackage{color}
\usepackage[pdf]{pstricks}
\usepackage{auto-pst-pdf}
\usepackage{pst-node}

\newcommand{\seq}{\mathrm{SEQ}}
\newcommand{\conceptsym}{N_{C}}
\newcommand{\rolesym}{N_{R}}
\newcommand{\constantsym}{N_{i}}
\newcommand{\Ot}{\mathcal{O}}

\newcommand{\Cn}{\mathrm{Cn}}
\newcommand{\V}{\mathcal{V}}
\newcommand{\Vcom}{\mathcal{V}}
\newcommand{\Vint}{\mathcal{V}'}

\newcommand{\con}[1]{\mathrm{#1}}

\newcommand{\I}{\mathcal{I}}

\newcommand{\BA}{\mathcal{B}}
\newcommand{\BAst}{\dot{\mathcal{B}}}
\newcommand{\ol}[1]{\overline{#1}}
\newcommand{\ars}{\mathrm{AR}}
\newcommand{\supp}{\mathrm{sp}}
\newcommand{\MRI}{\mathrm{MCS}}
\newcommand{\dremainder}{\rotatebox[origin=c]{270}{$\vdash$}}

\newcommand{\qed}{$\square$}

\newcommand{\false}{\bot}
\newcommand{\worevO}[2]{\mathbin{\mathopen{\otimes}^{#1}_{#2}}}
\newcommand{\orevO}[2]{\mathbin{\mathopen{\odot}^{#1}_{#2}}}
\newcommand{\nworevO}[1]{\mathbin{\mathopen{\otimes}^{#1}}}
\newcommand{\norevO}[1]{\mathbin{\mathopen{\odot}^{#1}}}
\newcommand{\orev}[1]{\mathbin{\mathopen{\boxdot}_{#1}}} 

\DeclareMathOperator{\conc}{\mathcal{C}}
\DeclareMathOperator{\pot}{Pow}
\DeclareMathOperator{\msc}{msc}

\newcommand{\ndoteq}{\not\doteq}
\DeclareMathOperator{\ub}{oa}
\DeclareMathOperator{\sel}{sel}
\newcommand{\worev}[1]{\mathbin{\mathopen{\otimes}_{#1}}} 
\newcommand{\gmorev}{\mathbin{\mathopen{\circ}}} 
\newcommand{\morev}[1]{\mathbin{\mathopen{\gmorev}_{#1}}}
\newcommand{\osel}[1]{\mathbin{\mathopen{\oplus}^{\sel}_{#1}}} 
\newcommand{\uk}{\sqsubseteq} 
\newcommand{\oder}{\sqcup} 
\newcommand{\und}{\sqcap}

\newtheorem{proposition}{Proposition}
\newtheorem{example}{Example}
\newtheorem{definition}{Definition}
\newtheorem{lemma}{Lemma}

\newenvironment{beweis}{\noindent{\bf Proof.} \hspace{0.5em}}{\hfill \qed\vspace{1ex}}
\newenvironment{postAllg}[1]{
\par\medskip\noindent
{\bf(#1)}\hspace{0.5em}}{\medskip}

\begin{document}

\title{Iterated Ontology Revision by Reinterpretation}

\author{\"Ozg\"ur L. \"Oz\c{c}ep\\Institute of Information Systems (IFIS)\\
University of L\"ubeck, Germany\\
oezcep@ifis.uni-luebeck.de}

\nocopyright
\maketitle

\begin{abstract}
Iterated applications of belief change operators are essential for different scenarios such as that of ontology evolution where new information is not presented at once but only in piecemeal fashion within a sequence. I discuss iterated applications of so called reinterpretation operators that trace  conflicts between ontologies back to the ambiguous of symbols and that provide conflict resolution strategies with  bridging axioms. The discussion centers on adaptations of the classical iteration postulates according to Darwiche and Pearl.  The main result of the paper is that  reinterpretation operators fulfill the postulates  for sequences containing only atomic triggers. For complex triggers,  a fulfillment is not guaranteed and indeed there are different  reasons for the different postulates why they should not be fulfilled in the particular scenario of ontology revision with well developed ontologies. 
\end{abstract}

\section{Introduction\label{sect:introduction}}
Iterated applications of belief change operators are essential for different scenarios such as that of ontology evolution where new information is not presented at once but only in piecemeal fashion within a sequence. Ontology evolution is a form of ontology change \cite{flouris08ontology} where an ontology modification is  triggered by changes in the domain or in the conceptualization. The response to the change is the application of (a set of) predefined operators. 

In this paper I consider the special scenario where an ontology (called the receiver's ontology) has to be changed along the arrival of a sequence of triggering bits of ontology  fragments coming from another ontology (sender's ontology). In the terminology of Flouris et al. \cite{flouris08ontology} the one-step change would be termed ontology merge as the purpose is to get a better understanding of the domain from merging two ontologies over the same domain. As in our setting the merge is directed I call the kind of change operation \emph{iterated ontology revision}.      

One instance of iterated ontology revision is given by iterated reinterpretation operators \cite{oezcep2010IGPL}. In these operators,  conflicts between the trigger and the receiver's ontology is explained by ambiguous use of terms. Consider an example of two online library systems with ontologies:  the sender may use $Article$ to denote publications in journals whereas the receiver may use $Article$ to denote publications in journals or proceeding volumes. As the ontologies are assumed to be over the same domain, the  receiver guesses relations on relations between her and the sender's uses and stipulates them as bridging axioms, e.g., stating that all articles in the sender's sense are articles in the receiver's sense. 

Now, a challenging aspect is to define adequateness criteria that iterated ontology revision operators should fulfill. I consider the classical iteration postulates of Darwiche and Pearl \cite{darwiche94logic} as possible candidates and state whether they are fulfilled by the reinterpretation operators. Moreover I  discuss, for each of them, whether it should be fulfilled at all. The main result of the paper is that  reinterpretation operators fulfill the postulates sequences with atomic triggers. For sequences of complex triggers,  a fulfillment is not guaranteed and indeed there are different reasons---corresponding to  different postulates---why they should not be fulfilled in the particular scenario of ontology revision with well developed ontologies.  

The rest of the paper is structured as follows. After some logical preliminaries and general terminology (Sect.\ \ref{sect:terminology})  the necessary definitions for reinterpretation operators are recapitulated (Sect.\ \ref{sect:ReinterpretationOperator}). Before the sections on related work and the conclusion, the adapted postulates for iterated revision, results on their fulfillment by reinterpretation operators,  and a discussion of the results are given in Sect.\  \ref{sect:postulatesIterated}.

\section{Terminology and Logical Preliminaries\label{sect:terminology}}
The reinterpretation framework described in the following works for any FOL theory but we consider here finite knowledge bases formulated in description logics. 

A non-logical DL vocabulary consists of concept symbols (= atomic  concepts) $\conceptsym$, role symbols $\rolesym$, and individual constants $\constantsym$. Using these, more complex concept descriptions can be built up in a recursive fashion. The set of possible concept constructors depends on the specific DL. We consider in particular the basic constructors $\sqcap, \sqcup, \neg, \exists$. $\conc(\V)$ is the set of all possible concept descriptions 
that can be built from the symbols in $\V$ in the given description logic. 

 The semantics is the usual Tarskian semantics based on interpretations $\I = (\Delta^{\I}, \cdot^{\I})$ with a domain $\Delta^{\I}$ and denotation function $\cdot^{\I}$ which gives for every $c \in \constantsym$ an element $c^{\I} \in \Delta^{\I}$, for every atomic concept $A \in \conceptsym$ a set $A^{\I} \subseteq \Delta^{\I}$ and for every role symbol $R \in \rolesym$ a binary relation $R^{\I} \subseteq \Delta^{\I} \times \Delta^{\I}$.  The denotation function is extended recursively to all concept descriptions in the usual manner.  For the ones we use here, we have $(C \und D)^{\I} = C^{\I} \cap D^{\I}$; $(C \oder  D)^{\I} = C^{\I} \cup D^{\I}$; $(\neg C)^{\I} = \Delta^{\I}\setminus C^{\I}$; $(\exists R. C)^{\I} = \{x \in \Delta^{\I} \mid \text{There is } y \in C^{\I} \text{ s.t. } (x,y) \in  R^{\I}\}$. Here $C,D \in \conc(\V)$  and  $R \in \rolesym$.  
 From concept descriptions one can built axioms which can be evaluated as true in (satisfied by) or false in (satisfied by)  an interpretation. We consider TBox (terminological Box)  axioms of the form 
 \begin{itemize}
 \item $C \uk D$ (concept subsumption) for $C,D \in \conc(\V)$  with semantics: $\I \models  C \uk D$ iff $C^{\I} \subseteq D^{\I}$;
 \item  $R_{1} \uk   R_{2}$ (role subsumption) for  $R_{1}, R_{2} \in \rolesym$ with semantics  $\I \models  R_{1} \uk R_{2}$ iff $(R_{1})^{\I} \subseteq (R_{2})^{\I}$
 \end{itemize}
 Moreover, we consider ABox axioms (assertional axioms) of the form $C(a)$ and $R(a,b)$ for $C \in \conc(\V)$, $a,b \in \constantsym$ and $R \in \rolesym$. The semantics is $\I \models C(a)$ iff $a^{\I} \in C^{\I}$ and $\I \models R(a,b)$ iff $(a,b) \in R^{\I}$. We call ABox axioms of the form $A(a)$ and $\neg A(a)$ with $A \in \constantsym$ concept assertions or concept-based literals. $\hat{A}$ stands for   $A$ or $\neg A$. Additionally,  equalities $a \doteq b$, $a,b \in \constantsym$  may be allowed.       
 
 Consistency (= satisfiability) of a set of axioms $X$ means that there is an interpretation $\I$ making all axioms in $X$ true, for short $\I \models X$. 
Entailment is defined as usual by $O_{1} \models O_{2}$ iff for all $\I$: If $\I \models O_{1}$, then $\I \models O_{2}$. A consequence operator $\Cn$ gives the set of all axioms following from a set: $\Cn(X) = \{ax \mid X \models ax\}$. If necessary, one can specify the vocabulary of the axioms: $\Cn^{\V}(X)$ is the set of axioms over $\V$ following from $X$. $X \equiv^{\V} Y$ is shorthand for $\Cn^{\V}(X) = \Cn^{\V}(Y)$. 
By $\V(O)$ we denote all non-logical symbols occurring in the set of axioms $O$. $\conc(O) = \conc(\V(O))$.  

The ontology notion of this paper slightly extends the one known from the semantic web and DL community---the extension relying on the distinction between an internal vocabulary $\V'$ and a public vocabulary $\V$: 
\begin{definition} An ontology $\Ot$ is a triple $\Ot = ( O, \V, \V' )$ consisting of a set of axioms $O$ over a logic with non-logical symbols $\V$ (the public vocabulary) and $\V'$ (the internal vocabulary). 
\end{definition}
In the following I will abuse terminology by calling also the set of axioms $O$ ontology. 

Let  $O_{1} \dremainder O_{2}$ be the \emph{dual remainder sets modulo $O_{2}$} \cite{delgrande08horn}. This is the set of inclusion maximal subsets $X$ of $O_{1}$ that are consistent with $O_{2}$, i.e., $X \in  O_{1} \dremainder O_{2}$ iff $X \subseteq O_{1}$, $X \cup O_{2}$ is consistent and for all $Y\subseteq O_{1}$ with $X \subsetneq Y$  the set $Y \cup O_{2}$ is not consistent.
  
We are going to deal with substitutions as means to realize name space dissociations. 
The set of  \emph{ambiguity compliant resolution substitutions}, denoted  $\ars(\V, \V')$,  
 consists of substitutions of symbols in $\V$ by symbols in $\V \cup \V'$.  
Here, we assume  $\V \cap \V' = \emptyset$ where $\V'$ is the set of symbols used for internalization.  The substitutions in $\ars(\V, \V')$   get as input a non-logical symbol in $\V$ (a constant, an atomic concept or role in DL speak)  and map it either to itself or to a new non-logical symbol (of the same type) in $\V'$.  The set of symbols $s \in \V$ for which $\sigma(s) \neq s$ is called the \emph{support} of $\sigma$ and is denoted $\supp(\sigma)$. In the following I use postfix notation for substitutions, i.e., $X\sigma = \sigma(X)$. Moreover,  I use the following  shorthands $\supp_{i}(\sigma) = \supp(\sigma) \cap \constantsym$ and $\supp_{CR}(\sigma) = \supp(\sigma) \cap (\conceptsym \cup \rolesym)$.   A substitution with support $S$ is also denoted by $\sigma_{S}$.  For  substitutions $\sigma_{1}, \sigma_{2} \in \ars(\V, \V')$ we define an ordering by: $\sigma_1 \leq \sigma_2$  iff $\supp(\sigma_1) \subseteq \supp(\sigma_2)$. 
 $\ars(\V, \V')$  can be partitioned into equivalence classes of substitutions that have the same support. We assume that for every equivalence class a representative substitution $\Phi(S) \in \ars(\V, \V')$	with support $S$ is fixed. $\Phi$ is called a \emph{disambiguation schema}.

\section{Reinterpretation Operators\label{sect:ReinterpretationOperator}}
This section recapitulates the definitions of ontology revision operators called reinterpretation operators \cite{oezcep2010IGPL,oezcep08towards}. 
The envisioned scenario is that of two agents holding well-developed ontologies, one called receiver's ontology, the other called sender's ontology. The ontologies are over the same domain and the receiver gets bits of information from the sender's ontology that she wants to integrate into her ontology in order to get a better, more fine-grained model of the domain. A challenging aspect is to preserve the consistency of the ontology. The kind of inconsistency that is considered here is that of inter-ontological ambiguity: the sender and the receiver may use the same symbol with different meanings (compare for example the different uses of $Article$ in the example below).   

 So, the conflict resolution strategy that is exploited by the reinterpretation operators is based on disambiguating symbols.  The sender or the receiver has to reinterpret  an ambiguous symbol. In the more interesting non-monotonic setting, that I consider in this paper, it is always the receiver who reinterprets the ambiguous symbol---by storing the old symbol in a new name space and relating her use of the symbols to the  sender's use by bridging axioms. This is in line with classical (prioritized) belief revision where one has full trust in the  trigger information.   In \cite{oezcep2010IGPL} these reinterpretation operators are called type-2 operators, contrasting them with type-1 operators in which it is the sender's terminology that is reinterpreted. 
 
The weak reinterpretation operators $\nworevO{}$ and strong reinterpretation operators $\norevO{}$ are binary operators with a finite set of ontology axioms ontology  as left and right argument. 
 The following example \cite{oezcep12minimality}   demonstrates the main ideas for the weak reinterpretation operators. 
\begin{example}\label{ex:motivationReint}
Consider the sets of ontology axioms $O_{1}$, $O_{2}$ of the receiver and sender, resp.:  
  \begin{eqnarray*}
  O_{1} &= &\{Article(pr_{1}), Article(pr_{2}), \neg Article(bo_1)\}\\
  O_{2} &=& \{\neg Article(pr_{1})\} 
  \end{eqnarray*}
   Applying the weak reinterpretation operator $\nworevO{}$ gives the following set of axioms: 
\begin{eqnarray*}
O_{1} \nworevO{} O_{2} &=& \{Article'(pr_{1}), Article'(pr_{2}),\\ 
&& \neg Article'(bo_1),  \neg Article(pr_{1}),\\
&& Article \uk Article'\}
 \end{eqnarray*}
 For the purpose of the example I assume that only concept and role symbols but not constant symbols may be used ambiguously. So,  the above conflict between the sender's and receiver's ontology can only be caused by different uses of the atomic concept $Article$. The receiver (holder of $O_{1}$) gives priority to the sender's use of $Article$ over her use of $Article$, and hence she adds $\neg Article(pr_{1})$ into the result $O_{1} \nworevO{} O_{2}$. The receiver's use of $Article$ is internalized, i.e., all occurrences of $Article$ in $O_{1}$  are substituted by a new symbol $Article'$. This step of internalization will also be called the step of dissociation or disambiguation. 
Additionally, the receiver adds hypotheses on the semantical relatedness (bridging axioms) of her and the sender's use of $Article$, here $Article \uk Article'$ which states that $Article$ is a subconcept of $Article'$.  
 \end{example}
Technically, the disambiguation is realized by uniform substitutions from   $\ars(\V, \V')$ (see section on logical preliminaries).  For the disambiguation,  
one has to deal with the a potential multiplicity of conflicts. The minimal conflict symbol sets defined below describe the smallest  sets of symbols which have to be disambiguated in order to resolve conflicts. 
 
 \begin{definition} For ontologies $O_{1}, O_{2}$ over $\V$  the set of \emph{minimal conflicting symbols sets}, $\MRI(O_{1}, O_{2})$,  is defined as follows:  
$$\begin{array}{l}
\MRI(O_{1}, O_{2}) =\\ 
\phantom{X}\{\;S \subseteq \V \mid \text{There is a }  \sigma_S \in \ars(\V, \V'), \text{ s.t. }\\
\phantom{X \{\;S \subseteq \V \mid} \text{$O_{1}\sigma_S \cup O_{2}$ is consistent,  and for }\\
\phantom{X \{\;S \subseteq \V \mid}\text{all $\sigma_{S_{1}} \in\ars(\V, \V') $} \text{ with } \sigma_{S_{1}} <  \sigma_S\\ 
\phantom{X \{\;S \subseteq \V \mid}   \text{$O_{1}\sigma_{S_{1}} \cup O_{2}$ is not consistent. } \}
\end{array}$$
\end{definition} 
Following the strategy of AGM partial meet revision \cite{agm1985}, we assume that a selection function $\gamma_{1}$ selects candidates from $\MRI(O_{1}, O_{2})$ to be used for the resolution: $\gamma_{1}(\MRI(O_{1}, O_{2})) \subseteq  \MRI(O_{1}, O_{2})$. So the symbol set defined by $S^{\#} = \bigcup \gamma_1(\MRI(O_{1}, O_{2}))$ is the set of symbols which will be internalized.   

To regain as much as possible from the receiver's ontology in the ontology revision result,  the disambiguated symbols of $S^{\#}$ are related by bridging axioms. Depending on what kind of bridging axioms are chosen, different revision operators result. In this paper we consider two classes of bridging axioms, the \emph{simple bridging axioms} and the \emph{strong bridging axioms}  \cite{oezcep08towards}.  Let  $\sigma = \sigma_{S} \in \ars(\V, \V')$ be a substitution with support  $S \subseteq \V$. 
Let  $P$ be a concept or role symbol  in $S$, $\sigma(P) = P'$.  
\begin{definition}
Let  $\sigma = \sigma_{S} \in \ars(\V, \V')$ for $S \subseteq \V \cap (\conceptsym \cup \rolesym)$. The set of \emph{simple bridging axioms}  w.r.t. $\sigma$ is 
 \begin{equation*}\BA(\sigma) = \{P \uk P', P' \uk P  \mid  P \in S\}\end{equation*}
 The set of \emph{strong bridging axioms}  w.r.t. $\sigma$ is defined  as:
$$ \begin{array}{l}
\BAst(\sigma, O) =\\
\phantom{X}  \{C\sigma \uk s \mid C \in \conc(O), s \in  \supp_{CR}(\sigma) , O \models C \uk s\} \cup\\
\phantom{X} \{s \uk C\sigma \mid C \in \conc(O), s \in  \supp_{CR}(\sigma), O \models  s \uk C\} \cup\\
\phantom{X} \{s \doteq s\sigma \mid s \in \supp_{i}(\sigma)\} 
\end{array}$$
\end{definition}

 In case of conflict, not all bridging axioms of $\BA(S^{\#})$ (resp. $\BAst(S^{\#})$) can be added to the  integration result (compare Ex. \ref{ex:motivationReint}). Hence, one searches for subsets that are compatible with the union of the internalized ontology and sender ontology, $O_{1}\sigma \cup O_{2}$.  That means, possible candidate sets of bridging axioms can be described by dual remainder sets (see section on logical preliminaries) as  $\BA(\sigma) \dremainder (O_{1}\sigma \cup O_{2})$.  Again, as there is no preference for one candidate over the other we assume that a second selection function $\gamma_{2}$ is given with $\gamma_{2}(\BA(\sigma) \dremainder (O_{1}\sigma \cup O_{2})) \subseteq (\BA(\sigma) \dremainder (O_{1}\sigma \cup O_{2})$. The intersections of the selected bridging axioms is the set of bridging axioms added to the integration result. (Compare this with the partial meet revision functions of AGM \cite{agm1985}).
 
  \begin{definition}\label{def:reinterpretationOperator} Let $\V, \V'$ be disjoint vocabularies and $\Phi$ a disambiguation scheme. Moreover let $\gamma_1, \gamma_2$ be selection functions and for short let $\ol{\gamma} = (\gamma_1, \gamma_2)$.  For any ontology $O_{1}$ and $O_{2}$ over $\V$ let  $S^{\#} = \bigcup \gamma_1(\MRI(O_{1}, O_{2}))$ and  $\sigma = \Phi(S^{\#})$. Then the \emph{weak reinterpretation operator} $\nworevO{\ol{\gamma}}$ and the \emph{strong reinterpretation operator} $ \norevO{\ol{\gamma}}$ are defined as follows:  
$$\begin{array}{rcl}
O_{1} \nworevO{\ol{\gamma}} O_{2} &=& O_{1}\sigma \cup O_{2} \cup \bigcap \gamma_2\big(\BA(\sigma) \dremainder (O_{1}\sigma \cup O_{2}) \big)\\
O_{1} \norevO{\ol{\gamma}} O_{2} &=& O_{1}\sigma \cup O_{2} \cup {}\\
&&\phantom{XXXX} \bigcap \gamma_2\big(\BAst(\sigma, O_{1}) \dremainder (O_{1}\sigma \cup O_{2}) \big)
\end{array}$$
 \end{definition}
 The definition for weak reinterpretation operators is the same as in \cite{oezcep08towards}, the definition of the strong operators is an extension.  

In the following I will simplify the discussion by simplifying the first step of internalization: In the internalization step now all symbols of the receiver are internalized. There is only a selection function for bridging axioms. Due to the fact that the maximal candidates of bridging axioms are used in the reinterpretation operators unnecessary internalizations  do not occur. I therefore can write in the following, e.g.,
  $\nworevO{\gamma}$ instead of $\nworevO{\ol{\gamma}}$.    
 
 A particularly interesting case of ontology change appears in the context of  ABox update  \cite{ahmeti14UpdateISWC,gutierrez11updating}, where the trigger informations are assertional axioms (ABox) axioms. I consider the special case that only atomic bits from the sender ontology  occur as trigger, namely,  the trigger $O_{2}$ is of the form $O_{2} = \{A(a)\}$ or of the form $O_{2} = \{\neg A(a)\}$.
  That is, the trigger is a concept assertion with an atomic symbol ($A \in \conceptsym$) or the negated atomic symbols $\neg A$.  
For this special case particular strong reinterpretation operators can be defined \cite{oezcep2010IGPL}.   
 The first class assumes that within the underlying DL a most specific concept w.r.t.\ an ontology exists. $C$ is a most specific concept for $b$ in the ontology $O$ iff $O \models C(b)$ and for all $C'$ s.t.\ $O \models C'(b)$ also $O \models C\sqsubseteq C'$. The most specific concept is unique modulo concept equivalence (w.r.t.\ $O$), hence it is denoted by $\msc_{O}(b)$. 
  
\begin{definition} \label{def:starkeReintOperatorenFuerTrigger}\index{Reinterpretationsoperator!starker (für Literale)}
Let $\Ot = (O, \Vcom, \Vint)$ be an ontology, $\Phi$ a disambiguation scheme,  $A \in \Vcom \cap \conceptsym$ and   $b \in \Vcom \cap \constantsym$. Assume $\sigma = [A/A']$ is the substitution fixed by  $\Phi$ and assume that  $\msc_{O}(b)$ exists.   The  \emph{msc-based strong reinterpretation operators for concept-based literals}  $ \orev{}$) are defined as follows:\\
If $O \cup \{ A(b)\} \not\models \bot$ let $O \orev{}  A(b) = O \cup \{ A(b)\}$. Else:  
$$\begin{array}{l}
O \orev{}  A(b)=\\
\phantom{X}  \sigma(O) \cup \{A(b), A' \uk  A,   A \uk  A' \oder \msc_{O\sigma}(b)\}
\end{array}$$
\noindent 
If $O \cup \{ \neg A(b)\}\not\models \bot $,  let $O \orev{}  \neg A(b) = O \cup \{ \neg A(b)\}$. Else: 
$$\begin{array}{l}
O \orev{}  \neg A(b) =\\ 
\phantom{X}\sigma(O) \cup \{\neg A(b), A \uk  A', A'  \uk  A \oder \msc_{O\sigma}(b)\}
\end{array}$$
\end{definition}  
The following examples illustrates the second case.  
\begin{example}
Assume that the receiver's ontology from the beginning is extended with two additional facts on the ``problematic'' entity $pr_{1}$:
\[O^+_{1} = O_{1} \cup \{publishedIn(pr_{1}, proc1), Proceed(proc1) \}\]
The most specific concept of $pr_{1}$ w.r.t. $O^+_{1}$ is 
 \[\msc_{O^+_{1}}(pr_{1}) = Article \sqcap \exists publishedIn. Proceed\]
 Hence the result of strong reinterpretation w.r.t.\ triggering concept assertions $O^+_{1} \orev{} \neg Article(pr_{1})$ adds the following additional bridging axiom    
\[Article' \uk Article \oder (Article' \sqcap \exists publishedIn. Proceed) \]
This says that the wider use of Article by the receiver adds (only) those publications in proceedings into the extension.  
\end{example}  
  
The selection-based strong operators for triggering literals provide  more bridging axioms between the internalized and non-internalized symbols.

\begin{definition} \label{def:selektionsBasierteReintOpFuerLiterale}  
Let $\Ot = (O, \Vcom, \Vint)$ be an ontology, $\Phi$ a disambiguation scheme,  $A \in \Vcom \cap \conceptsym$ and   $b \in \Vcom \cap \constantsym$. Assume $\sigma = [A/A']$ is the substitution fixed by  $\Phi$ and that  $\msc_{O}(b)$ exists.  Moreover, let $\sel$ be an arbitrary selection function, defined as $\sel(X) \subseteq X$.    The  \emph{ selection-based strong reinterpretation operators for concept-based literals}  $ \osel{}$) are defined as follows (using auxiliary definitions for the specific bridging axioms):
\begin{eqnarray*} 
\ub(O, A(b), K') &=&
\{A \uk A' \oder C \mid C \in \conc(\Vcom \cup \Vint),\\
&& \phantom{\{X}O\sigma \models C(b) \text{ and } A \notin \V(C) \}
\\
\ub(O, \neg A(b), A') &=&
\{A' \uk A \oder C \mid C \in \conc(\Vcom \cup \Vint),\\
&& \phantom{\{X} O\sigma \models C(b) \text{ and } A \notin \V(C) \}
\\
O \osel{}  \alpha &=&
\left\{ \begin{array}{l}
O \cup \{ \alpha \}  
\quad \text{if  $O \cup \{ \alpha\} \not\models  \bot$}\\
{O \worev{} \{\alpha\} \cup \sel(\ub(O, \alpha, A'))}\\
  \phantom{O \cup \{ \alpha \}  } \quad \text{else}
\end{array} \right.
\end{eqnarray*}
\end{definition}

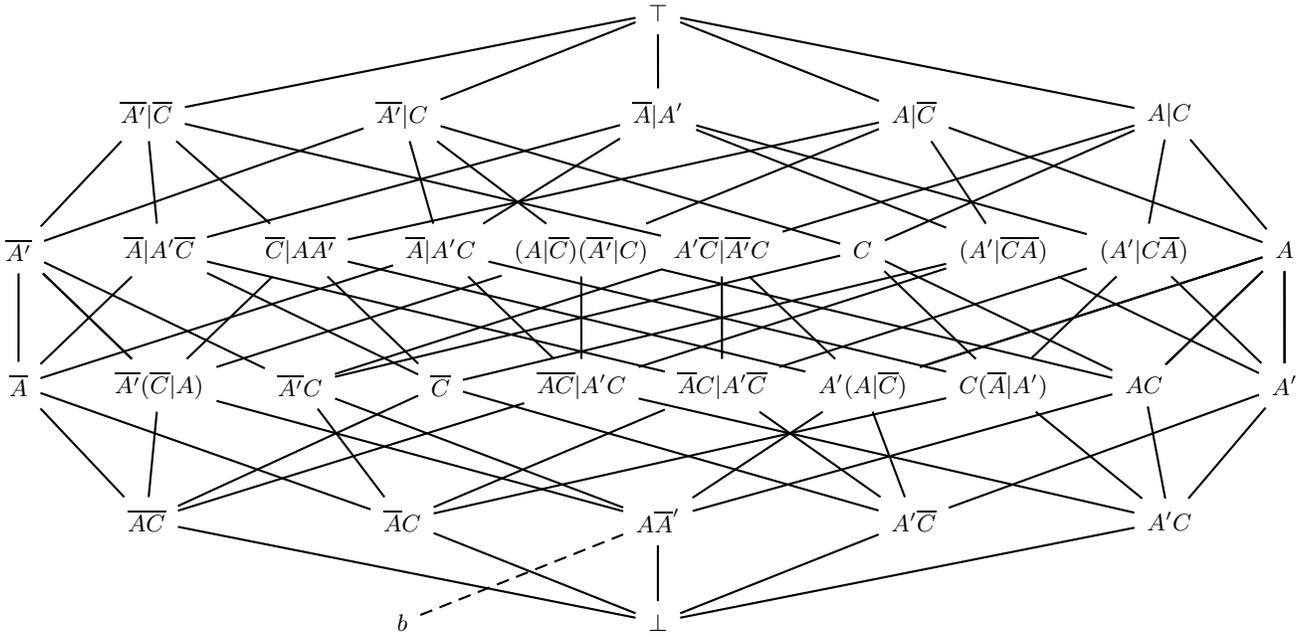
\begin{figure*}[t]
\begin{center}
  \psset{xunit= 1.7mm,yunit=0.9mm,runit=1mm,nodesep=1.7mm} 
\newlength{\y}
\setlength{\y}{5mm}
\begin{small}
\begin{pspicture}(0,0)(100,85)
$
\rput(50,90){\rnode{n11}{\top}}%
\rput(10, 75){\rnode{n21}{ \ol{A'}|\ol{C} }}
\rput(30, 75){\rnode{n22}{\ol{A'}|C}}
\rput(50, 75){\rnode{n23}{\ol{A}|A'}}
\rput(70, 75){\rnode{n24}{A|\ol{C}}}
\rput(90, 75){\rnode{n25}{A|C}}
\rput(0, 55){\rnode{n31}{\ol{A'}}}
\rput(11, 55){\rnode{n32}{\ol{A}|A'\ol{C}}}
\rput(22,55){\rnode{n33}{\ol{C}|A\ol{A'}}}
\rput(33, 55){\rnode{n34}{\ol{A}|A'C}}
\rput(44, 55){\rnode{n35}{(A|\ol{C})(\ol{A'}|C)}}
\rput(55, 55){\rnode{n36}{A'\ol{C}|\ol{A'}C }}
\rput(66, 55){\rnode{n37}{C}}
\rput(77, 55){\rnode{n38}{(A'|\ol{C}\ol{A})}}
\rput(88, 55){\rnode{n39}{(A'|C\ol{A})}}
\rput(99, 55){\rnode{n310}{A}}
\rput(0, 35){\rnode{n41}{\ol{A}}}
\rput(11, 35){\rnode{n42}{\ol{A'}(\ol{C}|A)}}
\rput(22, 35){\rnode{n43}{\ol{A'}C}}
\rput(33, 35){\rnode{n44}{\ol{C}}}
\rput(44,35){\rnode{n45}{\ol{A}\ol{C}|A'C}}
\rput(55, 35){\rnode{n46}{\ol{A}C|A'\ol{C}}}
\rput(66, 35){\rnode{n47}{A'(A|\ol{C})}}
\rput(77, 35){\rnode{n48}{C(\ol{A}|A')}}
\rput(88, 35){\rnode{n49}{AC}}
\rput(99, 35){\rnode{n410}{A'}}
\rput(10, 15){\rnode{n51}{\ol{A}  \ol{C}}}
\rput(30, 15){\rnode{n52}{\ol{A}  C}}
\rput(50, 15){\rnode{n53}{A  \ol{A}'}}
\rput(70, 15){\rnode{n54}{A'  \ol{C}}}
\rput(90, 15){\rnode{n55}{A' C}}
\rput(50,0){\rnode{n61}{\perp}}
\rput(30,0){\rnode{na}{b}}
$

\ncline[linestyle=dashed]{na}{n53}
%
\ncline{n11}{n21}
\ncline{n11}{n22}
\ncline{n11}{n23}
\ncline{n11}{n24}
\ncline{n11}{n25}
\ncline{n21}{n31}
\ncline{n21}{n32}
\ncline{n21}{n33}
\ncline{n21}{n36}
\ncline{n22}{n31}
\ncline{n22}{n34}
\ncline{n22}{n35}
\ncline{n22}{n37}
\ncline{n23}{n32}
\ncline{n23}{n34}
\ncline{n23}{n38}
\ncline{n23}{n39}
\ncline{n24}{n310}
\ncline{n24}{n38}
\ncline{n24}{n35}
\ncline{n24}{n33}
\ncline{n25}{n310}
\ncline{n25}{n39}
\ncline{n25}{n37}
\ncline{n25}{n36}
\ncline{n41}{n31}
\ncline{n41}{n32}
\ncline{n41}{n34}
\ncline{n42}{n31}
\ncline{n42}{n33}
\ncline{n42}{n35}
\ncline{n43}{n31}
\ncline{n43}{n36}
\ncline{n43}{n37}
\ncline{n44}{n32}
\ncline{n44}{n33}
\ncline{n44}{n38}
\ncline{n45}{n34}
\ncline{n45}{n35}
\ncline{n45}{n38}
\ncline{n46}{n32}
\ncline{n46}{n36}
\ncline{n46}{n39}
\ncline{n47}{n33}
\ncline{n47}{n36}
\ncline{n47}{n310}
\ncline{n48}{n34}
\ncline{n48}{n37}
\ncline{n48}{n39}
\ncline{n49}{n35}
\ncline{n49}{n37}
\ncline{n49}{n310}
\ncline{n410}{n38}
\ncline{n410}{n39}
\ncline{n410}{n310}

\ncline{n310}{n410}
\ncline{n310}{n49}
\ncline{n310}{n47}
\ncline{n42}{n31}

\ncline{n61}{n51}
\ncline{n61}{n52}
\ncline{n61}{n53}
\ncline{n61}{n54}
\ncline{n61}{n55}
\ncline{n51}{n41}
\ncline{n51}{n42}
\ncline{n51}{n44}
\ncline{n51}{n45}
\ncline{n52}{n41}
\ncline{n52}{n43}
\ncline{n52}{n46}
\ncline{n52}{n48}
\ncline{n53}{n42}
\ncline{n53}{n43}
\ncline{n53}{n47}
\ncline{n53}{n49}
\ncline{n54}{n44}
\ncline{n54}{n46}
\ncline{n54}{n47}
\ncline{n54}{n410}
\ncline{n55}{n45}
\ncline{n55}{n48}
\ncline{n55}{n49}
\ncline{n55}{n410}
%
\end{pspicture}
\end{small}
\end{center}
\caption{Concept lattice for $O \osel{} A(b)$ for the case  $O \models \neg A(b)$. We assume that the set of concepts chosen by  $\sel$ is representable as a concept  description $C$.  Here,  $A_1 \und A_2$ is abbreviated as  $A_1A_2$,   $\neg A$ by $\ol{A}$  and  $A_1 \oder A_2$ by $A_1|A_2$ }
\label{fig:BspSelektionsBasierteOperatoren}
\end{figure*}

Though the complexity of the trigger is low the induced concept lattice for the reinterpretation with $\osel{}$ is not trivial as illustrated by Fig. \ref{fig:BspSelektionsBasierteOperatoren}. Nonetheless, the figure does not suggest that the computation of the revision outcome is more complex than for other revision operators for DL ontologies: It just illustrates the subsumption connections of the concepts within the resulting ontology; the calculation of the lattice is not part of constructing the revision result.       

Without proof I state here some observations on the conservativity of  reinterpretation operators. Proofs can be found in \cite{oezcep2010IGPL}.   
\begin{proposition} \label{propRestrKonservativitaetTyp2}
Let $\Ot = (O, \Vcom, \Vint)$ be an ontology, $\Phi$ a disambiguation scheme,  $A \in \Vcom \cap \conceptsym$ and   $a,c \in \Vcom \cap \constantsym$. Assume $\sigma = [A/A']$ is the substitution fixed by  $\Phi$ and that $\msc_{O}(a)$ exists. Let $\alpha = A(a)$  $\epsilon = A(c)$  or $\alpha = \neg A(a)$ and $\epsilon = \neg A(c)$. Let $\beta$ be an assertion with  $\V(\beta) \subseteq (\Vcom\cup \V(O))\setminus\{A\}$. Let $\sel$ be a selection function for bridging axioms and  $\widetilde{\sel}$ a corresponding function selecting corresponding concepts:  $\widetilde{\sel}(\ub(O, \hat{A}(a),A')) = \{C\mid \hat{A} \uk \hat{A'} \oder C \in \sel(\ub(O,\hat{A}(a),A'))\}$.

 If  $O \models  \neg \alpha$, then 

$\begin{array}{rrlcl}
1.&
	 O \morev{} \alpha &\models \beta
	 &\mbox{ iff }& 
	 O \models \beta
	 \\
2.&
	O \morev{} \alpha &\models \epsilon 
	& \mbox{ iff }&
	 O \cup \{a \ndoteq c\} \models \epsilon 	
	 \\
3.&
	O \worev{} \alpha 	&\not\models  \neg\epsilon 
	\\
4.&
	O \orev{} \alpha 	&\models  \neg\epsilon
	&\mbox{ iff }&
	O \models  \neg\epsilon \mbox{ and }\\
	& & & &
	O \models \neg\msc_O(a)(c)
	\\	
5.&
	O \osel{} \alpha 	&\models  \neg\epsilon
	&\mbox{ iff }&
	O \models  \neg\epsilon \mbox{ and }\\
		& &   
	\multicolumn{3}{l}{O \models \neg \sqcap \widetilde{\sel}(\ub_2(O,\alpha,A'))_{[A'/A]}(c)	}
\end{array}$
\end{proposition}

\section{Postulates for Iterated Reinterpretation\label{sect:postulatesIterated}}
Many forms of ontology change \cite{flouris08ontology}, in particular ontology evolution \cite{kharlamov13capturing}, require the iterated application of a change operator under new bits of informations. In iterated belief revision, this problem is approached systematically by  defining, both, postulates and operators for iterated applications of  revision operators. The first systematic study of iterated belief revision goes back to the work of Darwiche and Pearl \cite{darwiche94logic} who stressed the  fact that the AGM postulates \cite{agm1985} are  silent w.r.t.\ the iterated application of operators. Indeed, the only postulates that can be said to touch some form of iteration are those dealing with the revision of conjunctions of triggers (supplementary postulates 7 and 8). 

I state those postulates in a form adapted to ontologies. 
\begin{postAllg}{RAGM 7}\label{postReintMrev7}
  $\Cn^{\Vcom}(O  \circ  (O_1 \cup O_2)) \subseteq \Cn^{\Vcom}((O \circ O_1) \cup O_2)$ 
\end{postAllg}

Postulate (RAGM 7) says that all sentences over $\Vcom$  following from  $O  \circ  (O_1 \cup O_2)$ are contained in the revision by $O_{1}$ followed by an expansion with   $O_2$. 

\begin{postAllg}{RAGM 8}\label{postReintMrev8}
If  $(O \circ O_1)  \cup O_2 \not\models \false$, then : 
\[ \Cn^{\Vcom}((O \circ O_1) \cup O_2) \subseteq \Cn^{\Vcom}(O  \circ  (O_1 \cup O_2))\]
\end{postAllg}

Postulate (RAGM 8) says that all sentences over $\Vcom$ following from the result of revising with $O_{1}$ and expanding with $O_{2}$ also follow from revising $O$ with the union of $O_{1}$ and $O_{2}$. A precondition is that  the revision result by   $O_1$ is compatible with $O_{2}$. 

In general, the reinterpretation operators do not fulfill these postulates. This can be shown with examples similar those provided by Delgrande and Schaub  \cite[p.~13]{delgrande03consistency}.  

But if one chooses particular selection functions for the reinterpretation operators on triggering ontologies, then  one can show that the postulates (RAGM 7) and (RAGM 8) are fulfilled.  This result is similar to an AGM theorem   \cite{agm1985}  which says that a partial meet revision operator on belief sets fulfills all  AGM postulates (in particular the supplementary ones) iff it can be defined as a  transitive relational  partial meet  revision operator. 

\begin{definition}
A selection function  $\gamma$ for bridging axioms is called  \emph{a maximum based selection function for bridging axioms  } iff the following holds:
\begin{enumerate}
\item $|\gamma(X)| = 1$ for all $\emptyset \neq X \subseteq \BA(\sigma_{\Vcom})$. 
\item  
 If $BA_1$ and $BA_2$ are non-empty sets of bridging  axioms from  $\BA(\sigma_{\Vcom})$, i.e.,  $BA_1, BA_2 \in \pot(\BA(\sigma_{\Vcom}))\setminus\{\emptyset\}$ s.t. for all  $X_2 \in BA_2$ there is a  $X_1 \in BA_1$ with $X_2 \subseteq X_1$,  and if additionally also   $\gamma(BA_1) \subseteq BA_2$ holds, then $\gamma(BA_2) = \gamma(BA_1)$. 
 \end{enumerate} 
 \end{definition}
Now one can show

 \begin{proposition}\label{beob:maximumsbasierteReintOpErfuellenErgPost} Let  $\Ot = ( O, \Vcom, \Vint )$, $\Ot_1 = ( O_1, \Vcom, \emptyset )$ and $\Ot_2 = ( O_2, \Vcom, \emptyset )$ be ontologies and  
 $\gamma$ be a maximum based selection function for bridging axioms. Then:\\
 If  $(O \nworevO{\gamma} O_1) \cup O_2$ is consistent, then : $O \nworevO{\gamma}{} (O_1 \cup O_2) = (O \nworevO{\gamma} O_1) \cup O_2$.
 \end{proposition}

This proposition shows that weak reinterpretation operators with maximum based selection function fulfill  (RAGM 8).  Moreover, one sees immediately that they fulfill the postulate  (RAGM 7) because if   $(O \nworevO{\gamma} O_1) \cup O_2$ is inconsistent, then trivially:   
\[\Cn^{\Vcom}(O  \nworevO{\gamma} (O_1 \cup O_2)) \subseteq \Cn^{\Vcom}((O \nworevO{\gamma} O_1) \cup O_2)\]

The supplementary postulates do not give constraints  for  the interesting case where for both triggers genuine revisions have to be applied. This motivated 
Darwiche and Pearl  \cite{darwiche94logic} to define four iteration postulates which, in an adaptation for the ontology revision scenario, are investigated in the following.    
 The results below show that the reinterpretation operators in general do not fulfill the postulates.

The postulates are, along the original ideas of Darwiche and Pearl  \cite{darwiche94logic}, described for finite sets of sentences (here: ontology axioms), and not for epistemic states as in their follow-up paper \cite{Darwiche97onthe}. Using the terminology of Freund and Lehmann \cite{freundlehmann94beliefRevision}, the type of iterated revision I consider in this paper is \emph{static}: There is no (used) encoding of the revision history in an epistemic state nor do I consider the dynamic change of revision operators from step to step. 
As we consider the justification of the iteration postulates not as a whole but one by one this approach does not stand in contradiction to the insights made in the follow-up paper by Darwiche and Pearl \cite{Darwiche97onthe}.

In the following postulates,  $\Ot = ( O, \Vcom, \Vint )$ is the initial ontology,  $\Ot_1 = ( O_1, \Vcom, \emptyset )$ the first triggering ontology and  $\Ot_2 = ( O_2, \Vcom, \emptyset )$ the second triggering ontology. Note that the trigger ontologies do not contain internal symbols---which fits the idea that only the public parts of the sender ontologies are communicated.

\begin{postAllg}{RDP 1} 
$\text{ If } O_2 \models O_1, \text{ then } (O \circ O_1) \circ  O_2 \equiv_{\Vcom} O \circ O_2$.
\end{postAllg}

In natural language: If the axioms of the second trigger ontology are stronger than the ones of the first trigger ontology, then the two-step outcome (relativized to the public vocabulary) is already covered by the revision  with the second trigger ontology. 
\begin{postAllg}{RDP 2}
$\text{If } O_1 \cup O_2 \text{ is not consistent,}\\  \phantom{X(  RDP 2 }\text{ then } (O \circ O_1) \circ O_2 \equiv_{\Vcom} O \circ O_2$.
\end{postAllg}

\noindent In natural language: If the axioms of the first and second trigger ontology are incompatible, then the two-step outcome (relativized to the public vocabulary) is already covered by the revision  with the second trigger ontology.  

\begin{postAllg}{RDP 3}
$\text{If } O \circ O_2 \models O_1, \text{ then } (O \circ O_1) \circ O_2 \models O_1$.
\end{postAllg}

\noindent  In natural language: If the revision by the second trigger ontology entails the first trigger ontology, then the entailment still holds for the revision with the first ontology followed by the second trigger ontology.

\begin{postAllg}{RDP 4}
$\text{If }  O_1 \cup (O \circ O_2) \text{ is consistent }\\ \phantom{c(RDP 4)} \text{ then so is } O_1 \cup (O \circ O_1) \circ O_2$.\\
In natural language: If the revision by the second trigger ontology is compatible with the first trigger ontology, then the compatibility still holds for the revision with the  first followed by the second trigger ontology. 
\end{postAllg}

As the following Proposition \ref{beob:DPPostulate}  shows, the fulfillment of all adapted iteration postulates cannot be guaranteed if the trigger is an ontology.  (This is the same as for the operators of Delgrande and Schaub \cite{delgrande03consistency}.)
 If the trigger is of atomic nature the situation is different due to the fact that there is only one symbol to be reinterpreted. For triggering literals only (RDP 2) is not fulfilled.

Proposition  \ref{beob:DPPostulate}  states  results for all reinterpretation operators mentioned in this paper:  Regarding the weak operators a distinction is made between triggering literals and ontologies.   
Table  \ref{tabelle:DPPostulate} summarizes the results. The rows contain the operators: The first three having concept-based literals as triggers, the last two ontologies.   The columns except for the last one refer to the iteration postulates. The last column gives a reference to the corresponding result in Proposition \ref{beob:DPPostulate}. 
Regarding the counterexamples I draw the following distinction---also reflected in the table: The weak counterexamples are those that construct ontologies for a specific selection function. The strong counterexamples are those that construct ontologies for any selection function.

 In the counter examples that were used to prove the negative results all reinterpretation operators reinterpret only atomic concepts and roles but not constants. As long as a conflict resolution by reinterpreting only concepts and roles is possible, the reinterpretation operators  can be modeled by a suitable definition of a  selection function $\gamma^{\con{CR}}$: $\gamma^{\con{CR}}$ selects only sets of bridge axioms that  contain  all identities for all constants which, in the end, means that the constants are not reinterpreted.  In this case I call $\gamma^{\con{CR}}$  a \emph{selection function that  prefers the reinterpretation of role and concept symbols}.

\begin{proposition}\label{beob:DPPostulate} Regarding the fulfillment of the adapted iteration postulates of Darwiche and Pearl \cite{Darwiche97onthe} the following results hold. 

\begin{enumerate}

\item\label{beobDPPostulateTyp2Literale} Reinterpretation operators for concept-based triggers ($\worev, \orev{}, \osel{}$) fulfill (RDP 1), (RDP 3)  and (RDP 4). \\
There are ontologies $\Ot, \Ot_1, \Ot_2$ such that  $\worev{}, \orev{}$  and $\osel{}$ (for all selection functions $\sel$) do not fulfill  (RDP 2).
There are ontologies  $\Ot, \Ot_1, \Ot_2$ and a selection function  $\sel$ such that  $\osel{}$ does not fulfill  (RDP 3) and does not fulfill (RDP 4).

\item\label{beobDPPostulateTyp2Ontologien}  For weak reinterpretation operators over triggering ontologies  $\worevO{\gamma}{}$ the following holds:\\ For all postulates (RDP $x$), $1 \leq x \leq 3$, there are ontologies  $\Ot, \Ot_1, \Ot_2$ such that for all selection functions  $\gamma^{\con{CR}}$ that prefer the reinterpretation of role and concept symbols  $\worevO{\gamma}{}$ does not fulfill (RDP $x$).\\
There are ontologies $\Ot, \Ot_1, \Ot_2$ and a selection function  $\gamma^{\con{CR}}$ that prefer the reinterpretation of role and concept symbols such that $\worevO{\gamma}{}$ does not fulfill (RDP 4).

\item\label{beobDPPostulateTyp2OntologienStark}  For strong reinterpretation operators over triggering ontologies   $\orevO{\gamma}{}$ the following holds:\\ For all postulates (RDP 1), (RDP 3), (RDP 4) there are ontologies  $\Ot, \Ot_1, \Ot_2$ and selection functions $\gamma$ such that 
$\norevO{\gamma}$ does not fulfill any of them.\\
There are ontologies $\Ot, \Ot_1, \Ot_2$ such that for all selection functions  $\gamma^{\con{CR}}$ that prefer the reinterpretation of concepts and role symbols  $\orevO{\gamma}{}$ does not fulfill (RDP 2).
\end{enumerate}
\end{proposition}

\begin{table*}[ht]
\begin{center}
\begin{tabular}{c|c|c|c|c|c}
Operator & (RDP 1) & (RDP 2) & (RDP 3) &(RDP 4) & Proposition\\
&&&&& \ref{beob:DPPostulate}.x\\\hline\hline
$\worev{}$ & + & -- & + & +& \ref{beobDPPostulateTyp2Literale}\\

$\osel{}$ & + & -- $(\forall \sel)$ & + & +\\

$\orev{}$ & + & -- & + & +\\\hline

$\worevO{\gamma}{}$ & -- ($\forall\gamma^{\con{CR}}$) & -- ($\forall\gamma^{\con{CR}}$) & -- ($\forall\gamma^{\con{CR}}$) & -- ($\exists\gamma^{\con{CR}}$) & \ref{beobDPPostulateTyp2Ontologien}\\\hline

$\orevO{\gamma}{}$ & -- ($\exists\gamma^{\con{CR}}$) & -- ($\forall\gamma^{\con{CR}}$) & -- ($\exists\gamma^{\con{CR}}$) & -- ($\exists\gamma^{\con{CR}}$) & \ref{beobDPPostulateTyp2OntologienStark}
\end{tabular}
\end{center}
\caption{Results of Proposition 
\ref{beob:DPPostulate} \newline A + entry  means that the postulate is fulfilled for all ontologies. A  -- entry means that there is an ontology such that   the postulate  is not fulfilled (only used for triggering literals). An entry of type  -- $(\forall \sel)$ resp.  {-- ($\forall \gamma$)} resp. 
 -- ($\forall \gamma^{\con{CR}}$) means that there are ontologies  s.t. for all selection functions $\sel$  resp. $\gamma$ resp.  $\gamma^{\con{CR}}$ the postulate is not fulfilled. An entry of type    -- $(\exists \gamma^{\con{CR}})$ means, that there is a selection function  $\gamma^{\con{CR}}$ such that the postulate is not fulfilled.%
 }
\label{tabelle:DPPostulate}
\end{table*}%

I discuss the outcomes of the proposition for the four postulates one by one starting with (RDP 2) which (in its original form given by Darwiche and Pearl) evoked most of the criticism.  
Regarding this postulate I follow the argument of Delgrande and Schaub \cite{delgrande03consistency} according to which (RDP 2) may make sense only for non-complex triggers. 
 For complex triggers, say in our case: complex ontologies $O_{1}$ (and $O_{2}$), does not work. Assume $O_{1}$ is made out of two sub-ontologies $O_{11}$ und $O_{12}$ s.t.  only  $O_{12}$ is not compatible with $O_{2}$. All those assertions that follow from $O \circ O_{1}$ on the basis of $O_{11}$ should be conserved after the revision with $O_{2}$.  But according to (RDP 2)  amnesic revision would be allowed if $O_{2}$ would not entail $O_{11}$: All sentences inferred with  $O_{11}$ would be eliminated in favor of the new ontology $O_{2}$. 

Regarding the first iteration postulate, the following simple example by Delgrande and Schaub \cite{delgrande03consistency} demonstrates its questionable status. Actually, for the proof of Proposition \ref{beob:DPPostulate}.\ref{beobDPPostulateTyp2Ontologien} I use adapted variants of this example.  
Consider the ontologies 
$O = \{\neg A(a)\}$, 
$O_1 = \{(A \oder B)(a)\}$ und 
$O_2 =  \{A(a)\} \models O_1$. 
Let $\gamma$ be a selection function that prefers the reinterpretation of concept and role symbols. 
The second ontology  $O_2$ ist stronger than the first ontology $O_1$.  Revision with $O_{2}$ leads to an ontology in which  $B(a)$ does not hold:  $O \nworevO{\gamma}{} O_2 \not\models B(a)$.    The revision with the first ontology leads to an ontology in which  $B(a)$ holds:  $O \nworevO{\gamma}{} O_1 = \{\neg A(a),  (A\oder B)(a)\} \models B(a)$.
The revision by the first and then by the second ontology   gives an ontology that still entails  $B(a)$: 
\begin{eqnarray*}
(O \nworevO{\gamma} O_1) \nworevO{\gamma} O_2 &=& \{\neg A(a), (A\oder B)(a), A'(a'),\\
&& \phantom{\{}A \uk A', a \doteq a', B \uk B',\\
&& \phantom{\{} B' \uk B\} \quad \models \quad  B(a)
\end{eqnarray*}
All preconditions in the antecedent of (RDP 1) are fulfilled but not the succedens:  $(O \nworevO{\gamma}{} O_1) \nworevO{\gamma}{} O_2 \not\equiv^{\Vcom} O \nworevO{\gamma}{} O_2$.   

There is no plausible revision operator for this particular ontology setting that would fulfill (RDP 1). 
Such an operator would have to fulfill $O \circ O_1 \models B(a) $ as  $O$ and $O_1$ are compatible. The revision with  $O_2$ should not eliminate  $B(a)$ as $B(a)$ is not relevant for the conflict: $(O \circ O_1) \circ O_2 \models B(a)$.  
Clearly, one could define syntax-sensitive revision operators on belief bases  s.t.  $(O \circ O_1) \circ O_2 \not\models B(a)$ so that the fulfillment of  (RDP 1) could be achieved also for this ontology configuration.  But this does not change the situation that also $(O \circ O_1) \circ O_2\models B(a)$ should be fulfilled. Moreover, syntax-sensitive belief-base operators are not appropriate for the revision of ontologies for which we would like to ensure (unique) syntax insensitive representations. So the only possibility for 
 $\circ$ to fulfill (RDP 1) is that  $O \circ O_2 \models B(a)$ holds. Such an operator $\circ$ that fulfills these conditions can be defined :  $O$ entails $(\neg A \oder B)(a)$ and $(\neg A \oder \neg B)(a)$.  If $\circ$ has a selection function  $\gamma$ that chooses  $(\neg A \oder B)(a)$, then  $O \circ O_2$ would entail $B(a)$.  But one could equally  have a selection function $\gamma'$ such that  $O \circ O_2 \not\models B(a)$ or even $O \circ O_2\models \neg B(a)$. There is no adequate reason for assuming that one has to prefer $\gamma$ over $\gamma'$.

The counter example against  (RDP 1) refers to triggering ontologies. For non-complex triggers such as concept-based literals a counter example cannot be constructed. Indeed: In this case all reinterpretation operators fulfill  (RDP 1) (besides (RDP 3) and (RDP 4) ( see Proposition \ref{beob:DPPostulate}.\ref{beobDPPostulateTyp2Literale}). 

Regarding the weak reinterpretation operator  for triggering ontologies $\worevO{\gamma}{}$ one can construct examples such that for all selection functions $\gamma$ that prefer the reinterpretation of concept and role symbols $\worevO{\gamma}{}$  does not fulfill the postulate (RDP 3)  (Proposition  \ref{beob:DPPostulate}.\ref{beobDPPostulateTyp2Ontologien}). For strong reinterpretation operators for triggering ontologies one can at least construct ontologies and at least one selection function showing the non-fulfillment (RDP 3). The counter example for the weak variants is based on the interplay of trivial revision (consistency case) and  non-trivial revision (inconsistency case): 
\begin{eqnarray*}
O &=& \{A(a), \neg B(a) \vee \neg A(c), A(b) \vee \neg A(e)\}\\
O_1 &=& \{\neg A(b)\}\\
O_2 &=& \{\neg A(a), B(a), A(c) \vee \neg A(b), A(e)\}
\end{eqnarray*}
$O$ and $O_1$ are chosen such that they are compatible and so  $O \nworevO{\gamma} O_1 = O \cup O_1 \models \neg A(e)$.  Due to the antecedens in  postulate (RDP 3) the revision by the second triggering ontology $O_{2}$ gives an ontology $O \nworevO{\gamma} O_2$ that entails the first trigger  $O_1$. The conflict resolution for  $O_2 \cup O$  is such that $O_1$ is not effected by it.    But a previous revision with  $O_1$ requires a different (additional)  conflict resolution with $O_2$ such that   $O_1$ is not entailed anymore: $(O \nworevO{\gamma} O_1) \nworevO{\gamma} O_2 \not\models \{\neg A(b)\} (= O_1)$. The reason that  $\neg A(b)$ cannot be inferred anymore is due to the fact that the conflict resolution  for $O \nworevO{\gamma} O_1$ and   $O_2$  leads to a reinterpretation of  $A$, and due to the fact that the  assertion  $\neg A(e)$, which follows from  $O \nworevO{\gamma} O_1$, has the same polarity as  $\neg A(b)$: namely, it is also negated.

This lost of  $\neg A(b)$ is due to the construction of the reinterpretation operators which implement a uniform reinterpretation: In case of conflicts all occurrences of symbols involved in the conflict are internalized. Only by introducing bridging axioms is it possible to regain assertions in the public vocabulary. But when the bridging  axioms are not expressive enough, then old sentences of the receiver  (such as  $\neg A(b)$ in this example) may not be entailed anymore.  
 This last discussion regarding (RDP 3) (and similarly for (RDP 4))  cannot be used as  general arguments against  (RDP 3) and (RDP 4) as adequate reinterpretation postulates.   One may construct plausible ontology revision operators fulfilling  (RDP 3) and (RDP 4), but these cannot implement uniform reinterpretation: They would have to do partial reinterpretation (as, e.g., done by Goeb and colleagues \cite{goeb07dynamic}).
 So, acceptable arguments  against (RDP 3) and (RDP 4) would have to support the requirement of uniformity within reinterpretation. And indeed,  there are good arguments in form of novel postulates that are motivated by typical requirements in ontology change settings: One wants to preserve the ontologies somehow in the ontology revision result and also wants them to be reconstructible. In particular these requirements occur when the ontologies are well-developed.   

I describe these postulates for the iterated scenario with a sequence $\seq$ of triggering ontologies. Let $\Ot = ( O, \Vcom, \Vint )$ be an ontology and let $\seq$ be a finite sequence of ontology axioms containing only symbols in the public vocabulary $\Vcom$.  

An operator $\gmorev$ that fulfills the iterated  preservation postulate  for the left argument (Preservation) has to guarantee that there is a substitution $\sigma$ s.t. the initial ontology $O$ is preserved in internalized form  $O\sigma$ in the result of iterated revision with a sequence $\seq$. 

\begin{postAllg}{Preservation} There is a substitution $\sigma$ s.t.: 
\[O\sigma \subseteq O \gmorev \seq\]
\end{postAllg}

An operator  $\gmorev$ that fulfills   (Reconstruction) has to guarantee the existence of a substitution $\rho$  such that the initial ontology $O$ and the set $set(\seq)$ of all triggering ontologies in the sequence $\seq$ are contained in a renamed variant of the revision result $(O \gmorev \seq)\rho$.  

\begin{postAllg}{Reconstruction} There is a substitution $\rho$ s.t.:  
\[O\cup set(\seq)\subseteq (O \gmorev \seq)\rho\]
\end{postAllg}

All reinterpretation operators of this paper fulfill both postulates. I state this proposition the proof of which is a slight adaptation of the proof given in \cite{oezcep2010IGPL} for triggering concept-based literals.   
\begin{proposition}\label{prop:ReintOperatorenErfuellenItPresRec}
Let  $\Ot = ( O, \Vcom, \Vint )$ be an ontology and  $\seq$ be finite sequence of setts of ontology axioms  over $\Vcom$ and $\gmorev$ a reinterpretation operator for triggering ontologies.  Then there exists  $\sigma$ and $\rho$, such that : 
\begin{enumerate}
\item $O\sigma \subseteq O \gmorev \seq$
\item $O\cup set(\seq)\subseteq (O \gmorev \seq)\rho$
\item For all symbols  $C \in \V(O) \cup \Vcom$ one has: $C = C\rho$. 
\end{enumerate}
\end{proposition}

\section{Related Work\label{sect:related}}
The  reinterpretation operators are constructed in a similar fashion as   those by Delgrande and Schaub \cite{delgrande03consistency} but differ in that they are defined not only for propositional logic but also for DLs (and FOLs). Moreover, I consider different stronger forms of bridging axioms than the implications of  \cite{delgrande03consistency}.    

Bridging axioms are special mappings that are used in the reinterpretation operator as auxiliary means to implement ontology revision. One may also consider  mappings by themselves as the objects of revision  \cite{qi09conflictBased,meilicke08reasoning}. A particularly interesting case of mapping revision comes into play with mappings used in the ontology based data access paradigm \cite{calvanese09ontologies}. These mappings are meant to lift data from relational DBs to the ontology level thereby mapping between close world of data and the open world of ontologies.  In this setting different forms of inconsistencies induced by the mappings can be defined (such as local vs. global inconsistency)  and based on this mapping evolution ensuring (one form of consistency) be investigated \cite{lembo15mapping}. 

In this paper I used reinterpretation operators as change operators on ontologies described in DLs. There are different other approaches that use the ideas of belief revision for different forms of ontology change such as ontology evolution over DL-Lite ontologies \cite{kharlamov13capturing} or ontology debugging  \cite{ribeiro09base}. As the consequence operator over DLs do not fulfill all preconditions assumed by AGM \cite{agm1985} one cannot directly transfer AGM constructions and ideas one-to-one to the DL setting as noted, e.g., by Flouris and colleagues  \cite{flouris05applying} and dealt in more depth for non-classical logics by Ribeiro \cite{ribeiro12belief}.     
 For the definition of the reinterpretation operators the constraint is not essential. Nonetheless, they lead to constraints in providing appropriate counter examples: namely ontologies expressible in the DL at hand.

\section{Conclusion\label{sect:conclusion}}
The paper discussed iterative applications of reinterpretation operators meant to handle conflicts due to ambiguous use of symbols in related and well-developed ontologies. Reinterpretation operators may also be used for solving consistencies not due to ambiguity but due to  false information---and indeed, the related revision operators in \cite{delgrande03consistency} do not talk about ambiguity. But reinterpretation operators cannot be used to solve inconsistencies that clearly cannot be explained by ambiguity: namely consistencies due to different constraints of the sender and the receiver regarding the number of possible objects in the domain (this was discussed under the  term \emph{reinterpretation compatibility} in \cite{oezcep08towards}).   

The reinterpretation for triggering  literals were shown to fulfill  (adapted versions of) classical  iteration postulates of Darwiche and Pearl \cite{darwiche94logic} whereas the reinterpretation operators for ontologies were shown in general not to fulfill them. Some of the postulates were criticized for general reasons.   Nonetheless, still one may consider other forms of reinterpretation operators that incorporate the reinterpretation history in order to define dynamic operators:     For example one might weight  symbols according to the number of times they were  reinterpreted and then use a comparison of the weights in the next iteration step in order to decided which symbols to reinterpret next.  

In addition to the general criticisms I discussed  the adequateness of the other postulates in view of the special ontology change scenario.  Here one cannot guarantee the fulfillment by reinterpretation operators that implement a uniform reinterpretation. But uniformity is necessary in order to guarantee the fulfillment of postulates that express the preservation and reconstructibility of the ontologies in the revision result.

\section*{Appendix: Proofs}
\subsection*{Proof of Proposition \ref{beob:maximumsbasierteReintOpErfuellenErgPost}}
For the proof we need the following lemma
 \begin{lemma}\label{lem:MonotoneErweiterungZweitesArgumentRestmenge}
 For all  $X_2 \in \BA(\sigma_{\Vcom}) \dremainder (O\sigma_{\Vcom} \cup O_1 \cup O_2)$ there is a  $X_1 \in \BA(\sigma_{\Vcom}) \dremainder (O\sigma_{\Vcom} \cup O_1)$ such that  $X_2 \subseteq X_1$.  
 \end{lemma}
 \begin{beweis} If $O\sigma_{\Vcom} \cup O_1 \cup O_2 \cup X_2$ is consistent, so is   $O_1 \cup O_2 \cup X_2$.  If $X_2$ is not maximal, then there is a superset  $X_1$ in $\BA(\sigma_{\Vcom}) \dremainder (O\sigma_{\Vcom} \cup O_1)$.  
 \end{beweis}
 
Define the following sets 
 \begin{eqnarray*}
 X_1 &=& \gamma(\BA(\sigma_{\Vcom}) \dremainder (O\sigma_{\Vcom} \cup O_1))\\
 X_2 &=&  \gamma(\BA(\sigma_{\Vcom}) \dremainder (O\sigma_{\Vcom} \cup O_1 \cup O_2))
 \end{eqnarray*} 
 With this notation we have  $(O \nworevO{\gamma} O_1) \cup O_2 = O\sigma_{\Vcom} \cup O_1 \cup O_2 \cup X_1$  and $(O \nworevO{\gamma} O_1) \cup O_2 = O\sigma_{\Vcom} \cup O_1 \cup O_2 \cup X_2$. 
Assume that  $(O \worevO{\gamma}{2} O_1) \cup O_2$ is consistent.  Then there is a  $X' \in  \BA(\sigma_{\Vcom}) \dremainder (O\sigma_{\Vcom} \cup O_1 \cup O_2)$ with  $X_1 \subseteq X'$. Because of  Lemma \ref{lem:MonotoneErweiterungZweitesArgumentRestmenge} there is a  $X'' \in \BA(\sigma_{\Vcom}) \dremainder (O\sigma_{\Vcom} \cup O_1)$ with $X' \subseteq X''$. Hence  $X_1 \subseteq X''$. But as  $X_1$ is inclusion maximal,  one gets  $X_1 = X'' = X'$ and hence  
 $X_1 \in \BA(\sigma_{\Vcom}) \dremainder (O\sigma_{\Vcom} \cup O_1 \cup O_2)$. 
Due to Lemma \ref{lem:MonotoneErweiterungZweitesArgumentRestmenge} the first precondition for maximum based selection functions was shown to hold. Now we showed also that the second condition holds, hence  $X_1 = X_2$. In particular: $O \nworevO{\gamma} (O_1 \cup O_2) = (O \nworevO{\gamma} O_1) \cup O_2$.

\subsection*{Proof of Proposition \ref{beob:DPPostulate}} 
All results hold trivially if  $O$ is not consistent. So for the following assume that $O$ is consistent.

\noindent \textbf{Proof of  \ref{beobDPPostulateTyp2Literale}.}  In the following let  $\morev{} \in \{\worev{}, \orev{}, \osel{}\}$. 

Proof for (RDP 1): As $O_1$ and $O_2$ are trigger literals,  $O_2 \models O_1$ holds iff $O_1 = O_2$. Hence $(O \morev{} O_1) \morev{}  O_2 =$ $(O \morev{} O_2) \morev{} O_2 =  O\morev{} O_2$, as the reinterpretation operators fulfill the success postulate.  

Counterexample for  (RDP 2): Let $O = \{A(b)\}$,  $O_1 =\{A(a)\}$ and $O_2 = \{\neg A(a)\}$. Then  $O_1 \cup O_2 \models \false$. On the one hand   $O \morev{} O_2 = \{A(b), \neg A(a)\} \models  A(b)$; on the other hand    $(O \morev{} O_1) \morev{} O_2 = \{A(b), A(a)\} \morev{2} \{\neg A(a)\}\not\models A(b)$ due to Proposition     \ref{propRestrKonservativitaetTyp2}.3 (for the weak  operators)  and Proposition \ref{propRestrKonservativitaetTyp2}.4 (for the strong operators) and again due to \ref{propRestrKonservativitaetTyp2}.3 for the selection based operators  $\osel{ }$, as in this case $(O \worev{} O_1) \worev{} O_2 \equiv (O \osel{} O_1) \osel{} O_2$.

Proof for (RDP 3): Let $O \morev{} O_2 \models O_1$.  

Case 1:  $O \cup O_2$ is consistent. As $O_1$ are  $O_2$ literals,  $O_2 \models O_1$ means that $O_1 = O_2$.  
Now  $(O \circ O_1) \morev{} O_2 = (O\morev{} O_2) \morev{} O_2 = O \cup O_2 = O \morev{} O_2$.

Case 2:  $O \cup O_2$ is inconsistent.
We show the proof for  $O_2 = \{A(a)\}$. (The case that $O_2 = \{\neg A(a)\}$ is proved similarly.)  Subcase 2.1: $A \notin \V(O_1)$.  Then from  $O \morev{2} O_2 \models O_1$ and    Proposition \ref{propRestrKonservativitaetTyp2}.1, it follows that  $O \models  O_1$, hence  $(O \morev{} O_1) = O \cup O_1$. Because of \ref{propRestrKonservativitaetTyp2}.1 one has: $(O \cup O_1) \morev{} O_2 \models O_1$.  Subcase  2.2: $A \in \V(O_1)$.  Let  $\morev{} = \worev{}$. Because of  $O \morev{} O_2 \models O_1$ and  Proposition \ref{propRestrKonservativitaetTyp2}.3 it must be the case that  $O_1$ contains  $A$ positively, i.e.,  $O_1 = \{A(c)\}$. Because $\worev{}$ fulfills the success postulate $O \worev{}\{A(c)\}  \models \{A(c)\}$ and hence also  $(O \worev{} \{A(c)\}) \cup \{a \ndoteq c\} \models \{A(c)\}$. Because of Proposition \ref{propRestrKonservativitaetTyp2}.2 one gets    $(O \worev{} \{A(c)\}) \worev{2} \{A(a)\} \models \{A(c)\}$.  Now consider the case  $\morev{} = \orev{}$. If $A$ is positive  in $O_1$, then  $O_1 = \{A(c)\}$. In this case again  Proposition \ref{propRestrKonservativitaetTyp2}.2  gives the result  $(O \orev{} O_1) \orev{} O_2 \models O_1$. In the other case $O_1$ is of the form  $O_1 = \{\neg A(c)\}$; using the assumption  $O \morev{} O_2 \models O_1$ one gets with Proposition  \ref{propRestrKonservativitaetTyp2}.4 that    $O \models \neg \msc_O (a)(c)$ and $O \models O_1$.  In particular  $O \orev{} O_1 = O \cup O_1 \equiv O$. Also  $\msc_{O \orev{} O_1}(a) = \msc_{O}(a)$. Hence from   $O \models \neg \msc_O (a)(c)$ we get  $O \orev{} O_1\models \neg \msc_{O\orev{} O_1}  (a)(c)$. 
   Because  $O \orev{} O_1 \models O_1$ holds, one can infer with  Proposition \ref{propRestrKonservativitaetTyp2}.4  that $(O \orev{} O_1) \orev{2} O_2 \models O_1$.  With a similar argument and  Proposition \ref{propRestrKonservativitaetTyp2}.5 one can show that the results holds for $\morev{} =  \osel{}$. 
 
Proof for  (RDP 4): Let $O_1 \cup (O \morev{} O_2)$ be consistent. 

Case 1: $O \cup O_2$ is consistent. Then $O \morev{} O_{2} = O \cup O_{2}$ and so $O_1 \cup (O \cup O_2) \not\models \bot$ due to assumption.  But then  $O \morev{} O_{1} = O \cup O_{1}$.

Case 2: $O \cup O_2$ is not consistent.  If also  $(O \morev{} O_1) \cup O_2$is consistent, then $(O \morev{} O_1) \morev{} O_2 = (O \morev{} O_1) \cup O_2$. As $O \morev{} O_1 \models O_1$ we then have  $(O \morev{} O_1) \morev{} O_2 \models O_1$, in particular $O_1 \cup (O \morev{} O_1) \morev{} O_2) \not\models \false$. Therefore  consider now the case that   $(O \morev{} O_1) \cup O_2$ is inconsistent. 

 We show the result for  $O_2 = \{A(a)\}$ (The argument for negative literals is similar).  Subcase 2.1: $A \notin \V(O_1)$. Due to success, $O \morev{} O_1 \models O_1$ and with Proposition  \ref{propRestrKonservativitaetTyp2}.1 it follows that   $(O \morev{} O_1) \morev{} O_2 \models O_1$. Because $(O \morev{} O_1) \morev{} O_2$ is consistent so is    $O_1 \cup (O \morev{} O_1) \morev{} O_2$. 
 Subcase 2.2: $A \in \V(O_1)$. Assume $O_1 \cup (O \morev{} O_1) \morev{} O_2 \models \bot$.  Is $O_1$ of form  $O_1 = \{\neg A(c)\}$, then due to Proposition \ref{propRestrKonservativitaetTyp2}.2 it holds that  $O_1 \cup O \morev{} O_1 \cup \{a \ndoteq c\}$ is not consistent. Because $O \morev{} O_1 \models O_1$ this can be the case only if $O  \morev{} O_1 \models a \doteq c$. Then   $O_1 = \{\neg A(a)\}$ which contradicts the assumption  $O_1 \cup (O \morev{} O_2) \not\models \false$. 

 If  $O_1$ has the form $O_1 = \{ A(c)\}$, then, due to  Proposition \ref{propRestrKonservativitaetTyp2}.3, this can only be the case if $\morev{} = \orev{}$ or  $\morev{} = \osel{}$. With the assumption that ($O_1 \cup (O \morev{} O_1) \morev{} O_2$ is not consistent, i.e. $(O \morev{} A(c)) \morev{} A(a) \models \neg A(c)$, one could infer with  Proposition \ref{propRestrKonservativitaetTyp2}.4 and  \ref{propRestrKonservativitaetTyp2}.5 that  $O \morev{} A(c) \models \neg A(c)$ which would mean that   $O_1 \cup (O \morev{} O_1)$ is not consistent----contradicting the consistency of   $O \orev{} O_1$ and the fact that $O \orev{2} O_1 \models O_1$.\\
\noindent \textbf{Proof of \ref{beobDPPostulateTyp2Ontologien}.}
Counter example for (RDP 1): Consider 
\begin{eqnarray*}
O &=& \{\neg A(a)\}\\
O_1 &=& \{(A \oder B)(a)\}\\
O_2 &=&  \{A(a)\}
\end{eqnarray*}
 Let be $\gamma$ an arbitrary selection function that prefers the reinterpretation of  role and concept symbols. Then we get on the one hand
$ O \nworevO{\gamma} O_2 = \{\neg A'(a'), A(a), A' \uk A, a \doteq a'\}$. 
And so $O \nworevO{\gamma} O_2 \not\models B(a)$. 
On the other hand 
$ O \nworevO{\gamma} O_1 = \{\neg A(a),  (A\oder B)(a)\} \models B(a)$. 
And last 
\begin{eqnarray*}
(O \nworevO{\gamma} O_1) \nworevO{\gamma} O_2 &=& \{\neg A'(a'), (A'\oder B')(a'), A(a),\\
&& \phantom{\{}A' \uk A, a \doteq a', B \uk B',\\
&& \phantom{\{}B' \uk B\}
 \end{eqnarray*}  
But then $(O \nworevO{\gamma} O_1) \models B(a)$,

Counter example for (RDP 2): See counter example for triggering literals.  

Counter example for (RDP 3): 
\begin{eqnarray*}
O &=& \{A(a), \exists R_1.A \uk \neg B, R_1(a,c), \exists R_2.A \uk A,\\
&& R_2(b,e)\}\\
O_1 &=& \{\neg A(b)\}\\
O_2 &=& \{\neg A(a), B(a), A(e), \exists R_3. A \uk A, R_3(c,b)\}
\end{eqnarray*}
For all selection functions $\gamma$ that prefer the reinterpretation of concept and role symbols one has: 
\begin{eqnarray*}
O \nworevO{\gamma} O_1 &=& O \cup O_1\\
O \nworevO{\gamma} O_2 &=&O\sigma_{\Vcom} \cup O_2 \cup \BA(\sigma_{\Vcom})\setminus\{A' \uk A \}\\
& \models &\neg A(b)\\
(O \nworevO{\gamma} O_1) \nworevO{\gamma} O_2&=& (O \cup O_1)\sigma_{\Vcom} \cup O_2 \cup {}\\
&& \BA(\sigma_{\Vcom})\setminus\{A' \uk A, A \uk A'\}\\ &\not\models& \neg A(b)
\end{eqnarray*}

Note that the  $O,O_1,O_2$ have simple structures and do not presuppose complex DL constructors.

Counter example for (RDP 4): Consider:
\begin{eqnarray*}
O &=& \{B(a), B(b) \vee C(b)\}\\
O_1 &=& \{\neg A(a), \neg B(b)\}\\
O_2 &=& \{\neg B(a) \vee A(a), \neg B(b), \neg C(b)\}
\end{eqnarray*}
Choose $\gamma$ such that the following results hold: 
\begin{eqnarray*}
O \worevO{\gamma}{2} O_2 &=& \{B'(a'), B'(b') \vee C'(b'),\\  &&\phantom{\{} \neg B(a) \vee A(a), \neg B(b), \neg C(b), \\
&&\phantom{\{}  a \doteq a', b \doteq b',\\
&& \phantom{\{} C \uk C', C' \uk C, B \uk B'\}\\
 &\not\models& \neg \bigwedge O_1
 \end{eqnarray*}
 \begin{eqnarray*}
O \worevO{\gamma}{2} O_1 &=& \{B(a), B(b) \vee C(b),\\
&& \neg A(a), \neg B(b) \} \models C(b)\\
(O \worevO{\gamma}{2} O_1) \worevO{\gamma}{2} O_2 &=& \{B'(a'), B'(b') \vee C'(b'),\\
&& \neg A'(a'), \neg B'(b'), \neg B \oder A(a),\\
&&   \neg B(b),\neg C(b), a \doteq a', b \doteq b',\\
&& B \uk B', B' \uk B, C \uk C',\\
 && A' \uk A\} \models \neg \bigwedge O_1
\end{eqnarray*} 
Note that we use here boolean ABoxes. As in the previous counterexample a purely DL counter example with standard ABoxes should be constructible. 

\noindent \textbf{Proof of  \ref{beobDPPostulateTyp2OntologienStark}.}  
Counter examples for  (RDP 1), (RDP 3) und  (RDP 4): Consider the same set of ontology axioms as in the corresponding counter examples for the weak reinterpretation  $\worevO{\gamma}{}$. The selection function $\gamma$ can be chosen such that the same results follow as in the case of  $\worevO{\gamma}{}$. 

Counter example for  (RDP 2): See counter example for  triggering literal.

\bibliographystyle{aaai}

\begin{thebibliography}{}

\bibitem[\protect\citeauthoryear{Ahmeti, Calvanese, and
  Polleres}{2014}]{ahmeti14UpdateISWC}
Ahmeti, A.; Calvanese, D.; and Polleres, A.
\newblock 2014.
\newblock Updating {RDFS} aboxes and tboxes in {SPARQL}.
\newblock In Mika, P.; Tudorache, T.; Bernstein, A.; Welty, C.; Knoblock,
  C.~A.; Vrandecic, D.; Groth, P.~T.; Noy, N.~F.; Janowicz, K.; and Goble,
  C.~A., eds., {\em The Semantic Web - {ISWC} 2014 - 13th International
  Semantic Web Conference, Riva del Garda, Italy, October 19-23, 2014.
  Proceedings, Part {I}}, volume 8796 of {\em Lecture Notes in Computer
  Science},  441--456.
\newblock Springer.

\bibitem[\protect\citeauthoryear{Alchourr\'{o}n, G\"{a}rdenfors, and
  Makinson}{1985}]{agm1985}
Alchourr\'{o}n, C.~E.; G\"{a}rdenfors, P.; and Makinson, D.
\newblock 1985.
\newblock On the logic of theory change: partial meet contraction and revision
  functions.
\newblock {\em Journal of Symbolic Logic} 50:510--530.

\bibitem[\protect\citeauthoryear{Calvanese \bgroup et al\mbox.\egroup
  }{2009}]{calvanese09ontologies}
Calvanese, D.; De~Giacomo, G.; Lembo, D.; Lenzerini, M.; Poggi, A.;
  Rodr\'{i}guez-Muro, M.; and Rosati, R.
\newblock 2009.
\newblock Ontologies and databases: The {DL-Lite} approach.
\newblock In Tessaris, S., and Franconi, E., eds., {\em Semantic Technologies
  for Informations Systems -- 5th Int. Reasoning Web Summer School (RW 2009)},
  volume 5689 of {\em Lecture Notes in Computer Science}. Springer.
\newblock  255--356.

\bibitem[\protect\citeauthoryear{Darwiche and Pearl}{1994}]{darwiche94logic}
Darwiche, A., and Pearl, J.
\newblock 1994.
\newblock On the logic of iterated belief revision.
\newblock In Fagin, R., ed., {\em Proceedings of the 5th Conference on
  Theoretical Aspects of Reasoning about Knowledge ({TARK}-94)},  5--23.

\bibitem[\protect\citeauthoryear{Darwiche and Pearl}{1997}]{Darwiche97onthe}
Darwiche, A., and Pearl, J.
\newblock 1997.
\newblock On the logic of iterated belief revision.
\newblock {\em Artificial intelligence} 89:1--29.

\bibitem[\protect\citeauthoryear{Delgrande and
  Schaub}{2003}]{delgrande03consistency}
Delgrande, J.~P., and Schaub, T.
\newblock 2003.
\newblock A consistency-based approach for belief change.
\newblock {\em Artificial Intelligence} 151(1--2):1--41.

\bibitem[\protect\citeauthoryear{Delgrande}{2008}]{delgrande08horn}
Delgrande, J.~P.
\newblock 2008.
\newblock Horn clause belief change: Contraction functions.
\newblock In Brewka, G., and Lang, J., eds., {\em Principles of Knowledge
  Representation and Reasoning: Proceedings of the 11th International
  Conference, {KR 2008}, Sydney, Australia, September 16-19, 2008},  156--165.
\newblock AAAI Press.

\bibitem[\protect\citeauthoryear{Eschenbach and
  {\"O}z\c{c}ep}{2010}]{oezcep2010IGPL}
Eschenbach, C., and {\"O}z\c{c}ep, {\"O}.~L.
\newblock 2010.
\newblock Ontology revision based on reinterpretation.
\newblock {\em Logic Journal of the IGPL} 18(4):579--616.
\newblock First published online August 12, 2009.

\bibitem[\protect\citeauthoryear{Flouris \bgroup et al\mbox.\egroup
  }{2008}]{flouris08ontology}
Flouris, G.; Manakanatas, D.; Kondylakis, H.; Plexousakis, D.; and Antoniou, G.
\newblock 2008.
\newblock Ontology change: classification and survey.
\newblock {\em The Knowledge Engineering Review} 23(2):117--152.

\bibitem[\protect\citeauthoryear{Flouris, Plexousakis, and
  Antoniou}{2005}]{flouris05applying}
Flouris, G.; Plexousakis, D.; and Antoniou, G.
\newblock 2005.
\newblock On applying the {AGM} theory to dls and {OWL}.
\newblock In Gil, Y.; Motta, E.; Benjamins, V.~R.; and Musen, M.~A., eds., {\em
  The Semantic Web - {ISWC} 2005, 4th International Semantic Web Conference,
  {ISWC} 2005, Galway, Ireland, November 6-10, 2005, Proceedings}, volume 3729
  of {\em Lecture Notes in Computer Science},  216--231.
\newblock Springer.

\bibitem[\protect\citeauthoryear{Freund and
  Lehmann}{2002}]{freundlehmann94beliefRevision}
Freund, M., and Lehmann, D.~J.
\newblock 2002.
\newblock Belief revision and rational inference.
\newblock {\em Computing Research Repository ({CoRR})} cs.AI/0204032.

\bibitem[\protect\citeauthoryear{Goeb \bgroup et al\mbox.\egroup
  }{2007}]{goeb07dynamic}
Goeb, M.; Reiss, P.; Schiemann, B.; and Schreiber, U.
\newblock 2007.
\newblock Dynamic {TBox}-handling in agent-agent-communication.
\newblock In Beierle, C., and Kern-Isberner, G., eds., {\em Dynamics of
  Knowledge and Belief. Proceedings of the Workshop at the 30th Annual German
  Conference on Artificial Intelligence ({KI}-2007)},  100--117.
\newblock Fernuniversit\"{a}t in Hagen.

\bibitem[\protect\citeauthoryear{Gutierrez, Hurtado, and
  Vaisman}{2011}]{gutierrez11updating}
Gutierrez, C.; Hurtado, C.; and Vaisman, A.
\newblock 2011.
\newblock Rdfs update: From theory to practice.
\newblock In {\em Proceedings of the 8th Extended Semantic Web Conference on
  The Semanic Web: Research and Applications - Volume Part II}, ESWC'11,
  93--107.
\newblock Berlin, Heidelberg: Springer-Verlag.

\bibitem[\protect\citeauthoryear{Kharlamov, Zheleznyakov, and
  Calvanese}{2013}]{kharlamov13capturing}
Kharlamov, E.; Zheleznyakov, D.; and Calvanese, D.
\newblock 2013.
\newblock Capturing model-based ontology evolution at the instance level: The
  case of dl-lite.
\newblock {\em J. Comput. Syst. Sci.} 79(6):835--872.

\bibitem[\protect\citeauthoryear{Lembo \bgroup et al\mbox.\egroup
  }{2015}]{lembo15mapping}
Lembo, D.; Mora, J.; Rosati, R.; Savo, D.~F.; and Thorstensen, E.
\newblock 2015.
\newblock Mapping analysis in ontology-based data access: Algorithms and
  complexity.
\newblock In et~al., M.~A., ed., {\em The Semantic Web - {ISWC} 2015 - 14th
  International Semantic Web Conference, Bethlehem, PA, USA, October 11-15,
  2015, Proceedings, Part {I}}, volume 9366 of {\em Lecture Notes in Computer
  Science},  217--234.
\newblock Springer.

\bibitem[\protect\citeauthoryear{Meilicke and
  Stuckenschmidt}{2009}]{meilicke08reasoning}
Meilicke, C., and Stuckenschmidt, H.
\newblock 2009.
\newblock Reasoning support for mapping revision.
\newblock {\em Journal of Logic and Computation}.

\bibitem[\protect\citeauthoryear{{\"O}z\c{c}ep}{2008}]{oezcep08towards}
{\"O}z\c{c}ep, {\"O}.~L.
\newblock 2008.
\newblock Towards principles for ontology integration.
\newblock In Eschenbach, C., and Gr{\"u}ninger, M., eds., {\em FOIS}, volume
  183,  137--150.
\newblock IOS Press.

\bibitem[\protect\citeauthoryear{{\"Oz\c{c}ep}}{2012}]{oezcep12minimality}
{\"Oz\c{c}ep}, {\"O}.~L.
\newblock 2012.
\newblock Minimality postulates for semantic integration.
\newblock In Konieczny, S., and Meyer, T., eds., {\em Proceedings of the
  workshop {BNC@ECAI2012}},  47--53.

\bibitem[\protect\citeauthoryear{Qi, Ji, and Haase}{2009}]{qi09conflictBased}
Qi, G.; Ji, Q.; and Haase, P.
\newblock 2009.
\newblock A conflict-based operator for mapping revision.
\newblock In et~al., B. C.~G., ed., {\em Proceedings of the 22nd International
  Workshop on Description Logics ({DL-09})}, volume 477 of {\em {CEUR} Workshop
  Proceedings}.

\bibitem[\protect\citeauthoryear{Ribeiro and Wassermann}{2009}]{ribeiro09base}
Ribeiro, M.~M., and Wassermann, R.
\newblock 2009.
\newblock Base revision for ontology debugging.
\newblock {\em J. Log. Comput.} 19(5):721--743.

\bibitem[\protect\citeauthoryear{Ribeiro}{2012}]{ribeiro12belief}
Ribeiro, M.
\newblock 2012.
\newblock {\em Belief Revision in Non-Classical Logics}.
\newblock SpringerBriefs in Computer Science. Springer.
\end{thebibliography}

\end{document}